\documentclass{article}
\usepackage{url}
\usepackage[utf8]{inputenc}
\usepackage[margin=1in]{geometry}
\usepackage[numbers,sort]{natbib}
\usepackage{hyperref}
\usepackage{csvsimple,booktabs,adjustbox}

\newcommand\contributionNote[1]{%
  \begingroup
  \renewcommand\thefootnote{}\footnote{\kern-5pt \textcolor{white}{\rule{5pt}{2ex}}#1}%
  \addtocounter{footnote}{-1}%
  \endgroup
}
\usepackage{bbm}

\def\hat{\widehat}

\usepackage{amsmath,amsfonts,bm}

\def\eqref#1{equation~\ref{#1}}
\def\1{\bm{1}}

\DeclareMathAlphabet{\mathsfit}{\encodingdefault}{\sfdefault}{m}{sl}
\SetMathAlphabet{\mathsfit}{bold}{\encodingdefault}{\sfdefault}{bx}{n}

\newcommand{\R}{\mathbb{R}}

\usepackage{url}
\usepackage[symbol]{footmisc}

\newcommand{\Real}{\mathbb{R}}
\newcommand{\N}{\mathcal{N}}

\usepackage{amsmath}
\usepackage{array}
\usepackage{booktabs}

\newcommand{\lTwoNorm}[1]{\left\lVert #1 \right\rVert_2}

\usepackage{graphicx}
\usepackage{wrapfig}

\begin{document}

\begin{center}

    \vspace*{0.5cm}
    \LARGE{Context-Scaling versus Task-Scaling in In-Context Learning}

\vspace*{0.5cm}

\large{Amirhesam Abedsoltan$^{1}$ \\ Adityanarayanan Radhakrishnan$^{2, 3}$ \\ 
Jingfeng Wu$^{5}$ \\ Mikhail Belkin$^{1, 4}$}

\vspace*{0.5cm}

\normalsize{$^{1}$Department of Computer Science and Engineering, UC San Diego} \\
\normalsize{$^{2}$Eric and Wendy Schmidt Center, Broad Institute of MIT and Harvard}\\
\normalsize{$^{3}$School of Engineering and Applied Sciences, Harvard University}\\
\normalsize{$^{4}$Halicioglu Data Science Institute, UC San Diego}\\
\normalsize{$^{5}$Simons Institute, UC Berkeley} \\
\end{center}

\begin{abstract}

Transformers exhibit In-Context Learning (ICL), where these models solve new tasks by using examples in the prompt without additional training.  In our work, we identify and analyze two key components of ICL: (1) context-scaling, where model performance improves as the number of in-context examples increases and (2) task-scaling, where model performance improves as the number of pre-training tasks increases.  While transformers are capable of both context-scaling and task-scaling, we empirically show that standard Multi-Layer Perceptrons (MLPs) with vectorized input are only capable of task-scaling.  To understand how transformers are capable of context-scaling, we first propose a significantly simplified transformer architecture without key, query, value weights. We show that it performs ICL comparably to the original GPT-2 model in various statistical learning tasks including linear regression, teacher-student settings.  Furthermore, a single block of our simplified transformer can be viewed as data dependent ``feature map'' followed by an MLP. This feature map on its own  is a powerful predictor that is capable of context-scaling but is not capable of task-scaling. We show empirically that concatenating the output of this feature map with vectorized data as an input to MLPs enables both context-scaling and task-scaling.  This finding provides a simple setting to study context and task-scaling for ICL.
\end{abstract}

\section{Introduction}

Pre-trained large language models exhibit In-Context Learning (ICL) capabilities, allowing them to adapt to new tasks based exclusively on input without updating the underlying model parameters \citep{brown2020language}.

\begin{table}[ht]
\centering
\begin{tabular}{>{\raggedright\arraybackslash}p{5.5cm} c >{\raggedright\arraybackslash}p{5.5cm}}
\toprule
\textbf{Input (Prompt)} & \textbf{Output} & \textbf{Task (Pattern)} \\
\midrule
$(1,2,3), (4,5,9), (10,-9,1), (5,6,?)$ & $11$ & In each triplet $(a,b,c)$: $c = a + b$ \\
\addlinespace
$(4,3,1), (9,0,9), (10,8,2), (17,17,?)$ & $0$ & In each triplet $(a,b,c)$: $c = a - b$ \\
\addlinespace
$(1,2,5), (2,3,8), (3,4,11), (5,6,?)$ & $17$ & In each triplet $(a,b,c)$: $c = a + 2b$ \\
\addlinespace
$(2,1,7), (3,4,18), (5,2,16), (4,3,?)$ & $17$ & In each triplet $(a,b,c)$: $c = 2a + 3b$ \\
\bottomrule
\end{tabular}
\label{tab:pattern_recognition}
\end{table}

In the table above, we provide an example of ICL. We prompt the Claude language model~\cite{Claude} using sequences of $N$ triples of numbers ($N=4$). In each row of the table, the first $N-1$ triples follow a given pattern. The model was able to infer the pattern and to fill in the missing number denoted by the question mark correctly. What makes this \textit{in-context learning} possible?

Recent research analyzed various ICL problems where, for example, the task data were generated using linear regression, student-teacher neural networks, and decision trees including~\cite{garg2022can,akyurek2022learning, bai2023transformers,ahn2023transformers,zhang2024trained,
raventos2023pretraining,wu2024many}.   In these problems, a transformer was first pre-trained on $T$ tasks where the data in each task was generated from a given family of functions. For example, each task may involve predicting the last element in a tuple as a linear combination of the other elements, as was shown in the table above.  The pre-trained transformer was then tested on $N$ samples from a new task that is drawn from the same general family but was not seen during pretraining. Such a setup allows for understanding the effect of various factors including model architecture, the number of pre-training tasks $T$, and context length $N$ on ICL.  

The ability of models to learn in context has been broadly defined as their ability to generalize to unseen tasks based on context examples without updating the model parameters.  We observe that there are two distinct aspects of generalization in ICL.  The first, which we call \textit{context-scaling}, refers to the ability of the model to improve as the context length $N$ increases while the number of pre-training tasks $T$ is fixed.  The second, \textit{task-scaling} refers to the ability of a model to improve as $T$ increases while $N$ is fixed.

\begin{figure}[ht]
    \centering
    \includegraphics[width=1\textwidth]{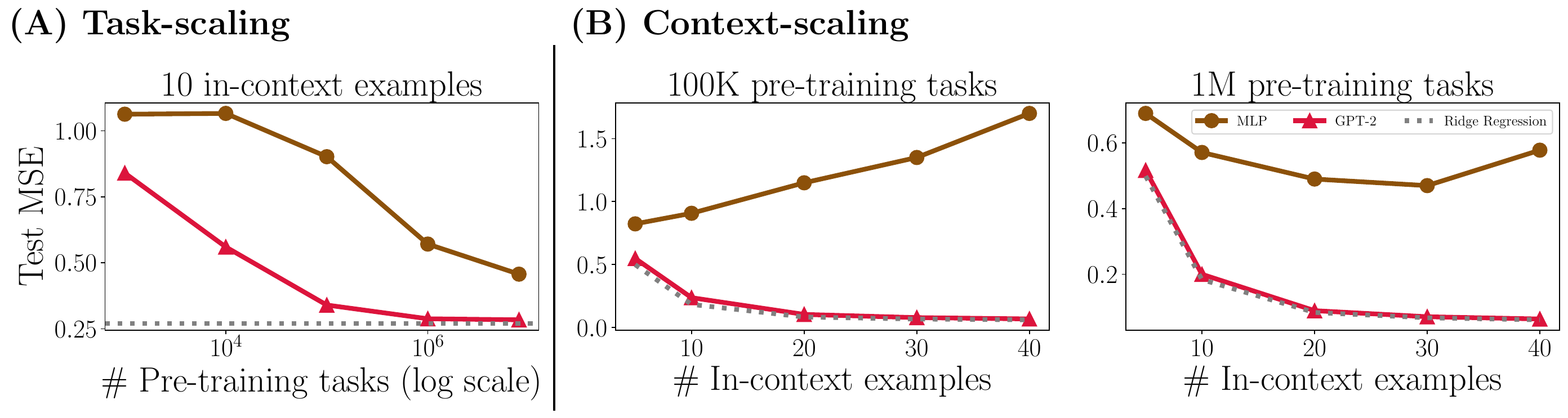}
    \caption{Task-scaling and context-scaling of GPT-2 architecture transformers versus MLPs for ICL with linear regression tasks.  \textbf{(A)} Task-scaling abilities of these models with $10$ in-context examples. \textbf{(B)} Context-scaling abilities of these models with $10^5$ (left) and $10^6$ (right) pre-training tasks. Experimental details are provided in Appendix \ref{app:A}.}
    
    % Transformer demonstrates both context and task scaling capabilities, while MLP with vectorized input achieves only task scaling. Detailed experimental setup and results are provided in Appendix \ref{app:E}. 
    % }

    \label{fig:weak_icl}
\end{figure}

It is a priori unclear whether a model capable of context-scaling is also capable of task-scaling and vice-versa. For example, as we show in Figure~\ref{fig:weak_icl}A,  both transformers and standard Multi-Layer Perceptrons (MLPs) are capable of task-scaling on ICL of linear regression tasks. In contrast, only transformers benefit from an increasing number of context examples as shown in Figure~\ref{fig:weak_icl}B. This raises the question:

\begin{center}
    \textit{What mechanism enables models such as  transformers, but not MLPs, to context-scale?}
\end{center}

% \textit{What mechanism enables transformers to context-scale but not MLPs?}

To identify such a mechanism, we begin by constructing a bare-bones transformer with all key, query, and value weight matrices fixed to be the identity matrix. We refer to our simplified model as Simplified GPT (SGPT). Despite its simplicity, we find that SGPT is competitive with GPT-2 architecture transformers \citep{radford2019language} for a variety of ICL problems considered in \citep{garg2022can} including linear regression, student-teacher networks, decision trees, and sparse linear regression.

Furthermore, we find that one block of SGPT applies a data dependent feature map that enables context-scaling.  To illustrate how such a feature map can be effective for context-scaling, consider the following input data for ICL:
\begin{align}
A = \begin{bmatrix}
x_1 & x_2 & \cdots &x_{N-1} &x_N  \\
y_1 & y_2 & \cdots &y_{N-1} &0
\end{bmatrix}^\top \in \mathbb{R}^{N\times(d+1)}~;
\end{align}
where $x_i \in \Real^d$ and our goal is to predict $y_N$. SGPT first applies a feature map $\psi:\mathbb{R}^{N\times(d+1)}\rightarrow\mathbb{R}^{N\times(d+1)}$ to $A$ and then trains an MLP on the last row of $\psi(A)$, denoted $\psi(A)_{N, :}$.

By varying $\psi$, we show that the scalar $\psi(A)_{N, d+1}$ itself is an effective estimate for $y_N$.  
For example, when $\psi(A) = (AA^\top) A$, $\psi(A)_{N, d+1}$ implements one-step of Gradient Descent (GD) for linear regression (using context examples $(x_i,y_i)_{i=1}^{N-1}$) \citep{von2022transformers}. When the data follow an isotropic Gaussian distribution, this estimator is consistent (as both the numbers of pre-training tasks and context examples grow) and nearly matches the optimally tuned ridge regression \citep{mahankali2023one,wu2024many}.
Furthermore, we show that $\psi(A)_{N, d+1}$ can implement the Hilbert estimate~\citep{devroye1998hilbert}, which provides a statistically consistent estimate for general families of tasks beyond linear regression as the context length $N$ approaches infinity. As such, our results provably establish that one attention layer alone is capable of context-scaling for any family of tasks.

Further, we empirically show that these features can enhance the capabilities of MLPs for context-scaling. We concatenate features from $\psi(A)_{N, :}$ with the vectorized input, $$A_v :=\begin{bmatrix}
    x_1^T& y_1 &x_2^T& y_2& \ldots &x_N^T& 0
\end{bmatrix}^T,$$ and provide the combined input $[A_v^T,~\psi(A)_{N,:}^T]$ to an MLP. We show that the resulting model exhibits both context-scaling and task-scaling.
%  which previously could only task-scale when trained on $A_v$. 

 We summarize our findings as follows:

\begin{itemize}

    \item We identify two regimes for characterizing generalization in ICL: \textit{context-scaling} and \textit{task-scaling}.  We observe that MLPs task-scale but do not context-scale in contrast to GPT2 architecture transformers, which exhibit both context and task-scaling.  
    
    \item We propose a simplified transformer, SGPT, with all key, query, and value weight matrices set to the identity matrix, exhibits both context and task-scaling and is competitive with GPT-2 on a range of ICL tasks.

    \item We analyze a one-layer version of SGPT, demonstrating that this model is capable of context-scaling solely through the use of a feature map, $\psi$, applied to input data.  We show that $\psi$ can be selected to perform kernel smoothing to impute the missing element in each task using the other in-context examples. Choosing $\psi$ based on the Hilbert estimate~\citep{devroye1998hilbert} results in a statistically optimal (consistent) estimate as the context length approaches infinity.

    \item We show empirically that concatenating the output of $\psi$ with vectorized data as an input to MLPs enables both context-scaling and task-scaling.
\end{itemize}

\section{Prior work}

\paragraph{ICL in controlled settings.}
The work by \cite{garg2022can} initiated the study of ICL in statistical learning tasks, such as linear regression, decision tree learning, and teacher-student neural network learning. They showed that transformers such as GPT-2 \citep{radford2019language}, when pre-trained with a large number of independent tasks (around $3.2\times 10^7$ independent tasks, each presented once), learn in-context.  Specifically, during inference time, the performance of pre-trained transformers empirically matches that of standard, specialized algorithms for these tasks.  These results were later extended to various other settings see, e.g., \citep{akyurek2022learning,raventos2023pretraining,bai2023transformers,li2023transformers,tong2024mlps}. In particular, \cite{raventos2023pretraining} empirically showed transformers can achieve nearly optimal ICL when pre-trained with multiple passes over a smaller, fixed training set of independent tasks. \cite{bai2023transformers} constructed transformers that implement optimal algorithms in context even if the tasks are generated from a mixture of distributions. In a recent work, \cite{tong2024mlps} empirically showed that pre-trained MLPs can achieve ICL when the context length is fixed during pre-training and inference. However, it was not known whether MLPs can achieve context-scaling, and our work empirically gives a negative answer.

\paragraph{Linear attention.} Most theoretical analyses of ICL has been in the setting of linear regression for fixed context lengths.  Specifically, \cite{von2022transformers} showed by construction that single linear attention can implement one-step gradient descent (GD) in context.  \cite{ahn2023transformers,zhang2024trained,mahankali2023one,zhang2024context} proved that the predictor given by pre-trained transformers with single linear layer of attention is equivalent to that given by one-step GD. \cite{wu2024many} proved the predictor from one-step of GD achieves the rate of optimally-tuned ridge regression. In addition, they show that the predictor given by one step of GD can be pre-trained to achieve this rate with finite independent tasks.
Later works such as \cite{cheng2024transformers} connected nonlinear attention to functional one-step GD in feature space. These works together have substantially furthered our understanding of single-layer linear attention transformers for ICL of linear regression with a fixed context length.
However, these results are primarily concerned with fixed context length settings and do not address the context-scaling abilities of pre-trained transformers.  Recent work~\cite{lu2024asymptotic} provided an asymptotic theory of ICL for linear attention transformers.  Unlike our analyses, they focused on settings in which the number of unique pre-training tasks (referred to as ``diversity'') scaled proportionally with the data dimension.

\paragraph{Softmax attention and kernel smoothers.}
The connection between softmax attention and kernel smoothers was first pointed out by \cite{tsai2019transformer}.
Specifically, by setting the query and key matrices to be the same, an attention head can be viewed as a kernel-smoother with a learnable positive semi-definite kernel. Empirical evidence suggests that using a shared matrix for query and key matrices does not significantly impact the performance of transformers \cite{tsai2019transformer,yu2024white}.
Later, theoretical works \cite{chen2024training,collins2024context} utilized this connection to study the ICL of softmax attention in both linear \citep{chen2024training} and nonlinear regression tasks \citep{collins2024context}.  In these settings, softmax attention can perform ICL by implementing a kernel smoother with a bandwidth parameter obtained by training query and key matrices. Compared with these works, we demonstrate that transformers can perform ICL using an attention component that does not contain trainable parameters (that is, setting query, key, and value matrices to identity matrices). Our results suggest that the transformer architecture is capable of performing ICL without needing to explicitly learn any hyperparameters in the attention head. We explain this by connecting to a hyper-parameter-free yet statistically consistent kernel smoother given by the Hilbert estimate \citep{devroye1998hilbert}.

\paragraph{Approximation ability of transformers.} 
The transformer is known to be a versatile architecture that can implement efficient algorithms (by forwarding passing an input prompt) in many scenarios  \citep{akyurek2022learning,bai2023transformers,guo2023transformers,lin2023transformers,gatmiry2024can}.
These results make use of query, key, and value weight matrices for constructing ICL algorithms. In this work, we show that these elements of transformers are not needed for building and training models that are competitive with GPT-2 architecture transformers for ICL tasks considered in prior works~\citep{garg2022can}.

\section{Preliminaries}\label{sec:prelim}

In this section, we outline the problem setup, training details, architectural details, and mathematical preliminaries for our work.

\paragraph{Problem formulation.} For all ICL tasks studied in our work, we consider $T$ pre-training tasks, each with input data of the form:
\begin{align}\label{eq:input_structure}
    A_t = \begin{bmatrix}
x_1^t & x_2^t & \cdots & x_N^t  \\
y_1^t & y_2^t & \cdots & y_N^t
\end{bmatrix}^\top \in \mathbb{R}^{(N)\times(d+1)},
\end{align}
where $t = 1, \ldots, T$ indexes the tasks, $N$ denotes the maximum context length, and $d$ denotes the input data dimension. 
We define the loss function as:
$$L(\theta; A_1, \ldots, A_T) := \frac{1}{T}\sum_{t=1}^T \left[\frac{1}{N}\sum_{i=1}^{N}(M_{\theta}(A_t^i)-y_{i+1}^t)^2\right],$$
where
$$A_t^i = \begin{bmatrix}
x_1^t & x_2^t & \cdots & x_i^t & x_{i+1}^t \\
y_1^t & y_2^t & \cdots & y_i^t & 0
\end{bmatrix}^\top \in \Real^{(i+1)\times(d+1)},$$ 
and $M_{\theta}(\cdot)$ denotes the model with trainable parameters $\theta$.  The tasks $A_t$ are uniformly sampled from the family of tasks $\mathcal{F}$ (e.g., linear regression with a Gaussian prior), representing the distribution of tasks relevant to the in-context learning problem.

\paragraph{Attention.}
Given three matrices $A_1, A_2, A_3 \in \mathbb{R}^{N \times m}$, attention layers implement functions $g:\Real^{N \times m} \times \Real^{N \times m} \times \Real^{N \times m} \to \Real^{N\times m}$ defined as follows,
\begin{align}\label{eq:attentionhead_gpt2}
    g(A_1, A_2, A_3) := \phi\left(\frac{1}{\sqrt{m}} A_1 A_2^\top\right) A_3;
\end{align}
where $\phi:\mathbb{R}^{N\times N} \to \mathbb{R}^{N\times N} $ is a generic function that could be a row-wise softmax function \citep{vaswani2017attention}, an entry-wise activation function such as ReLU, or just an identify map.  For self-attention layers, we are typically given one input matrix $A \in \mathbb{R}^{N \times m}$ and three weight matrices $W_Q, W_K, W_V \in \mathbb{R}^{m \times m}$.  In this case, the matrices $A_1, A_2, A_3$ are computed respectively as $A W_Q$, $A W_K$, and $A W_V$.

\paragraph{Kernel functions.}\label{par:notation_kernel}  Kernel functions are positive-semidefinite functions that map pairs of inputs to real values \citep{scholkopf2002learning}.  Formally, given $x,y \in \mathbb{R}^d$, a kernel $K: \mathbb{R}^d \times \mathbb{R}^d \rightarrow \mathbb{R}$ is a function of the form $K(x,y) = \langle\psi(x), \psi(y)\rangle_{\mathcal{H}}$, where $\psi: \mathbb{R}^d \rightarrow \mathcal{H}$ is referred to as a feature map from $\mathbb{R}^{d}$ to a Hilbert space $\mathcal{H}$.  For matrix inputs $A \in \mathbb{R}^{m \times d}$ and $B \in \mathbb{R}^{n \times d}$, we let $K(A,B) \in \mathbb{R}^{m \times n}$ such that $K(A,B)_{ij} = K(A_i, B_j)$ where $A_i$ and $B_j$ denote the $i$\textsuperscript{th} and $j$\textsuperscript{th} rows of $A$ and $B$ respectively.

\paragraph{Kernel smoother.} Given a kernel $K:\mathbb{R}^d \times \mathbb{R}^d \to \mathbb{R}$ and a set of points $(\mathbf{x}_i, y_i)_{i=1}^n$ where $\mathbf{x}_i \in \mathbb{R}^d$ and $y_i\in\mathbb{R}$, the kernel smoother estimate at point $\mathbf{x}$ is a function of the form
\begin{align}\label{def:kernel_smoother}
    \hat{f}_{K, n}(\mathbf{x}) := \frac{\sum_{i=1}^n K(\mathbf{x}, \mathbf{x}_i) y_i}{\sum_{i=1}^n K(\mathbf{x}, \mathbf{x}_i)}~.
\end{align}
We will reference kernel smoothers in Section \ref{sec:NTK_smoothers}.

\paragraph{Hilbert estimate.} The Hilbert estimate is a kernel smoother using the kernel
\begin{align}\label{eq:hilbert_kernel}
    H(x,x') := \frac{1}{\lTwoNorm{x-x'}^d},
\end{align}
where $\lTwoNorm{\cdot}$ denotes the $\ell_2$-norm in $\Real^d$.  The key property of the Hilbert estimate that we use is that it is a asymptotically optimal (consistent) estimate.  In particular, at almost all $x$, as the number of samples $n$ goes to infinity, $\hat{f}_{H, n} \to f^*$ in probability where $f^*(x) = \mathbb{E}[y | X = x]$ denotes the optimal predictor~\citep{devroye1998hilbert}.

\section{Simplified transformer model performs ICL}\label{sec:exps}

In this section, we introduce our simplified transformer model, SGPT, and demonstrate that it is competitive with GPT-2-type architectures on various ICL tasks. To construct SGPT, we fix all key, query, and value weights to be the identity matrix in the GPT-2 architecture. Consequently, the attention mechanism (defined in~\eqref{eq:attentionhead_gpt2}) reduces to the function: \begin{align} \label{eq:psi}
g(H) := \phi(HH^\top) H \in \Real^{N\times (d+1)}. \end{align}

We define $\phi$ to be a function that performs row-wise $\ell_1$ normalization on its argument. To further simplify our model, we remove the final linear layer of each MLP block (i.e., our MLP blocks have one linear layer), along with batch normalization and the skip connection after the MLP layer. These details are further outlined in Appendix \ref{app:A}.

We consider the following ICL tasks from prior work: (1) linear regression with a single noise level \citep{akyurek2022learning}, (2) linear regression with multiple noise levels used in \citep{bai2023transformers}, (3) sparse linear functions used in \citep{garg2022can}, (4) two-layer ReLU neural networks \citep{garg2022can}, and (5) decision trees \citep{garg2022can}. Below, we explain the problem setup and state our results for each of these five synthetic tasks.

\paragraph{Linear regression with fixed noise level.} The problem setting is as follows:
\begin{gather*}
x \in \R^d \sim \N(0, I_d), \quad y = \beta^\top x + \epsilon ~  
\text{with } \beta \sim \N\left(0, \frac{I_d}{d}\right), \; \epsilon \sim \N(0, \sigma^2).
\end{gather*}
\begin{figure}[ht]
    \centering
    \includegraphics[width=0.8\textwidth]{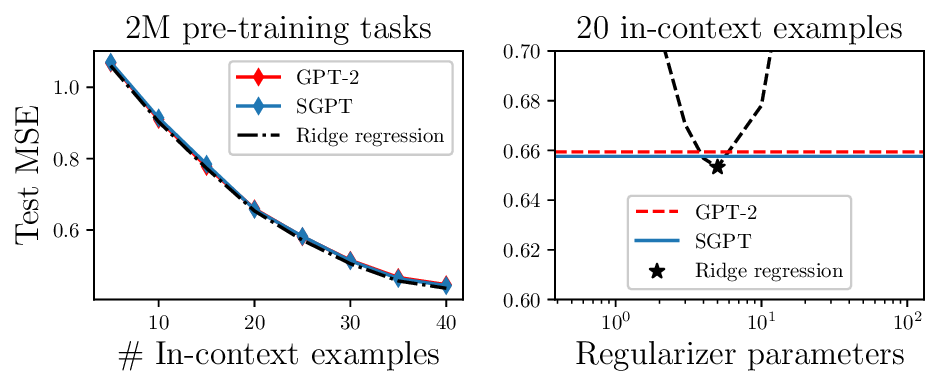}
    \caption{Linear regression with a single noise level. \textbf{Left panel.} Performance across varying context lengths (context-scaling). \textbf{Right panel.} Effect of regularization on performance for a fixed number of in-context examples. Experimental details are given in Appendix \ref{app:A}.}
    \label{fig:linear_icl_fixed_noise}
\end{figure}
In this setting, prior work by \cite{bai2023transformers}, showed that on all context lengths, GPT-2 architecture transformers can perform comparably to task-specific, optimally-tuned ridge regression. In Figure \ref{fig:linear_icl_fixed_noise}, we provide evidence that SGPT matches the performance of these GPT-2 models.

\paragraph{Linear regression with multiple noise levels.} The problem setting is as follows:
\begin{gather*}
x \in \R^d \sim \N(0, I_d), \quad y_i = \beta^\top x_i + \epsilon \\
\text{with } \beta \sim \N\left(0, \frac{I_d}{d}\right) \text{ and } \epsilon \sim 
\begin{cases}
\N(0, \sigma_1^2), & \text{with probability } \frac{1}{2} \\
\N(0, \sigma_2^2), & \text{with probability } \frac{1}{2}
\end{cases}.
\end{gather*}

\begin{figure}[ht]
    \centering
    \includegraphics[width=\textwidth]{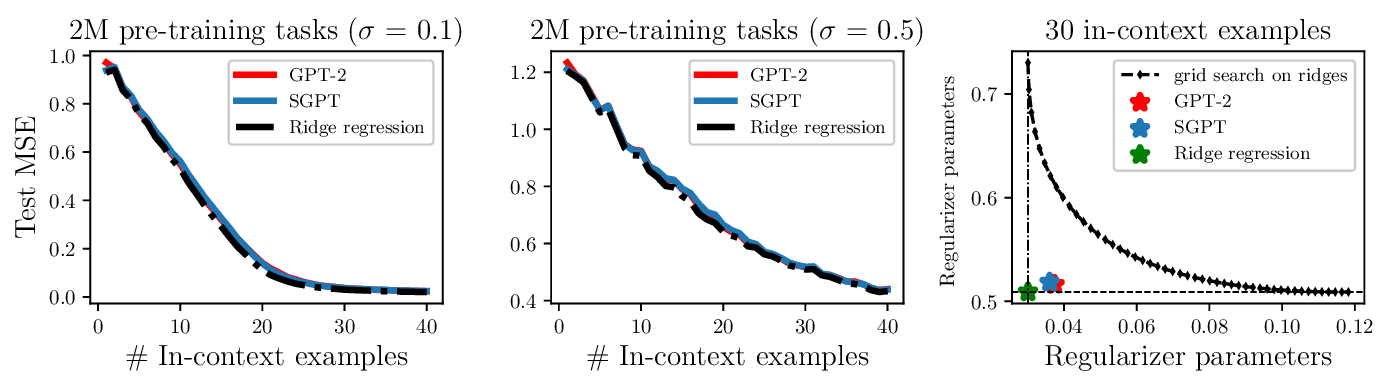}
    \caption{ Linear regression with multiple noise levels.    
    \textbf{Left and middle panels:} Performance across varying context lengths (context-scaling).
    \textbf{Right panel:} 
    Effect of regularization on performance for a fixed number of in-context examples. Experimental details are given in Appendix \ref{app:A}.
    }
    \label{fig:linear_icl_mixed_noise}
\end{figure}

In this setting, prior work by \cite{bai2023transformers} demonstrated that GPT-2 architecture transformers can achieve performance comparable to that of task-specific, optimally tuned ridge regression across all context lengths and for both noise levels. They refer to the model's ability to adapt to the noise level as \textit{algorithm selection}. In Figure~\ref{fig:linear_icl_mixed_noise}, We demonstrate that SGPT performs comparably to GPT-2 architecture transformers in this setting, exhibiting similar algorithm selection capabilities.

\begin{figure}[ht]
    \centering
    \includegraphics[width=\textwidth]{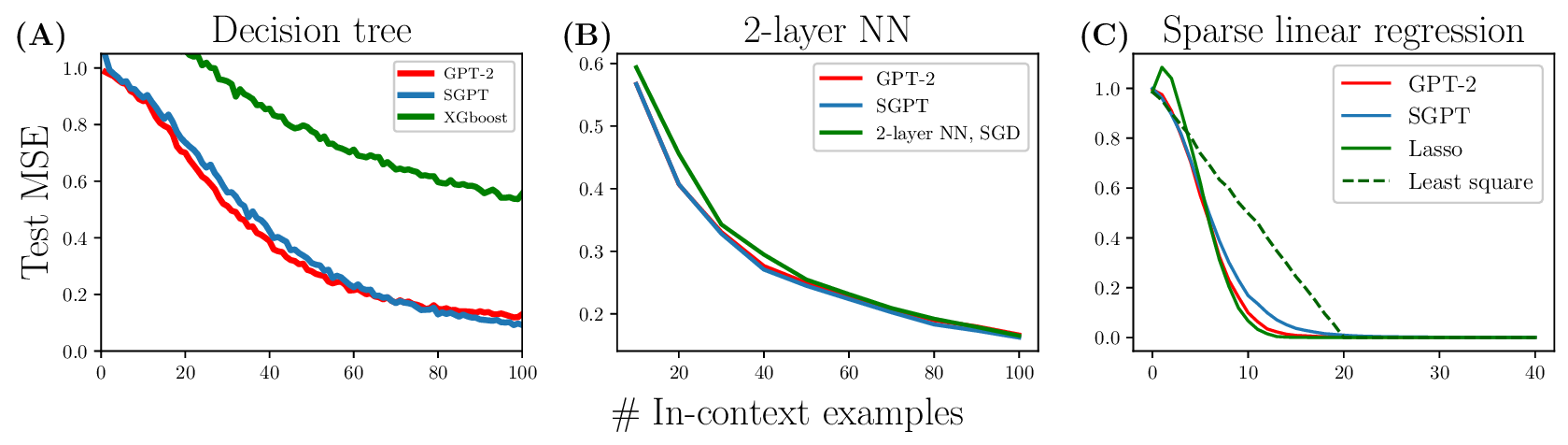}
     \caption{ Nonlinear ICL tasks. Context-scaling capability of SGPT versus GPT-2 architecture transformers when trained on 2 million pre-training tasks. In all cases, the errors are normalized so that the trivial zero predictor achieves an error of 1.$^*$ Experimental details are given in Appendix \ref{app:A}. 
     }

    \label{fig:nonlinear_icl}
\end{figure}

\paragraph{Two-layer ReLU Neural Networks.} Following the work of \cite{garg2022can}, we consider the following nonlinear problem setting where data for each task are generated using two-layer neural networks. In particular, data are generated according to
\begin{gather}\label{task:nn2layer}
x \in \mathbb{R}^d \sim \mathcal{N}(0, I_d), \quad y =  \sum_{j=1}^r \alpha_j \phi(w_j^\top x);
\notag
\end{gather}
where $\alpha_j, w_j$ are randomly initialized parameters of a fixed two-layer neural network, $\phi$ denotes the element-wise ReLU activation function, and $r=100,d =20$ (as selected in prior work). The work~\cite{garg2022can} demonstrated that GPT-2 architecture transformers can match the performance of student networks (i.e., networks of the same architecture initialized differently and trained using Adam optimizer~\cite{kingma2014adam}). In Figure \ref{fig:nonlinear_icl}(B), we show that SGPT can match the performance of GPT-2 architecture transformers on this task. 

\paragraph{Decision Tree.} 
Following the work of \cite{garg2022can}, we consider a nonlinear problem setting where data for each task are generated using depth-four trees. For a task corresponding to a tree $f$, we have:
\begin{gather}
x \sim \mathcal{N}(0, I_d), \quad y = f(x),
\end{gather}
Previously, \cite{garg2022can} demonstrated that GPT-2 architecture transformers can perform in-context learning on this family of non-linear functions, outperforming XGBoost as a baseline. In Figure \ref{fig:nonlinear_icl}(A), we show that SGPT is also capable of in-context learning (ICL) in this setting, performing comparably to GPT-2 architecture and similarly outperforming XGBoost. We trained XGBoost models using the same hyperparameters as in \citep{garg2022can}.

\paragraph{Sparse linear functions.} Following the work of~\cite{garg2022can}, we consider the class sparse linear regression problems. In this setting, data are generated according to  
\begin{gather}
x \in \R^d \sim \N(0, I_d), \quad y = \beta^\top x,
\end{gather}
where $\beta \sim \N\left(0, I_d\right)$ and we zero out all but $s$ coordinates of $\beta$ uniformly at random for each task.  As in prior work, we select $d=20, s=3$. In Figure \ref{fig:nonlinear_icl}C
, we demonstrate that SGPT is capable of ICL for this class of functions, performing comparably to GPT-2 architecture transformers and closely to the Lasso estimator~\citep{tibshirani1996regression}, while significantly outperforming the ordinary least square (OLS) baseline. 

 \footnotetext[1]{For decision trees, we found that GPT-2 performs poorly when using the input structure of concatenating $x$ and $y$. Therefore, we used the pre-trained model from\cite{garg2022can}}

Thus far, we have demonstrated SGPT is comparable to GPT-2 across various ICL tasks. In Appendix~\ref{app:D}, we revisit the experiment introduced in the introduction and show that GPT-2 and SGPT are capable of both context and task-scaling.

\section{Kernel smoothing can perform context-scaling}\label{sec:NTK_smoothers}

\begin{figure}[th]
    \centering
    \includegraphics[width=1\textwidth]{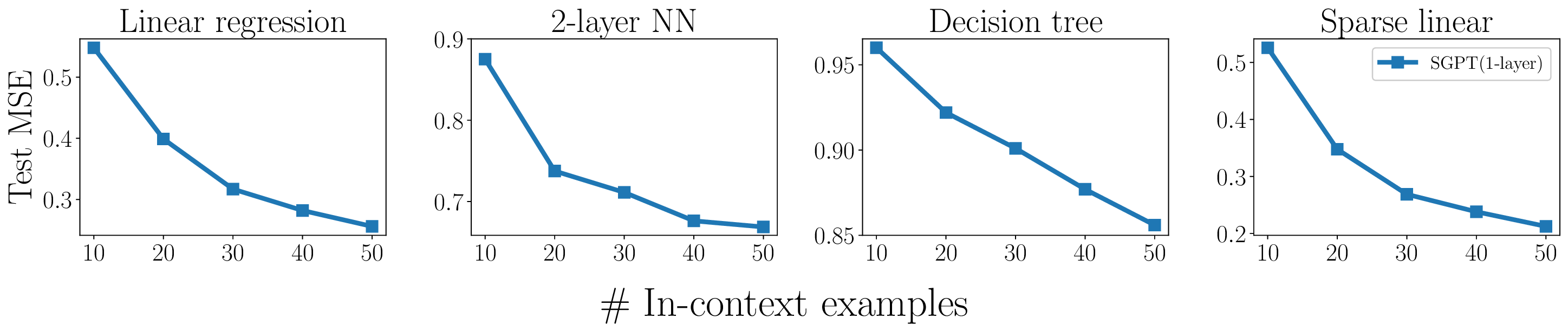}
    \caption{Context-scaling with one-layer SGPT.  Experimental details are provided in Appendix \ref{app:A}.}
    \label{fig:multi_context}
\end{figure}

We begin this section by demonstrating that even one layer of SGPT is capable of context scaling. In particular, in Figure~\ref{fig:multi_context}, we train a one-layer model on five different context lengths and test on the same lengths for the tasks considered in the previous section.  In all four problem settings, it is evident that using more context improves performance. Below, we analyze this simplified one layer model in order to pinpoint how it is capable of context scaling.

In particular, the model we analyze is identical to one layer of SGPT up to the omission of remaining skip connections (for more details, see Appendix \ref{app:A}).  The model implements a function $f: \mathbb{R}^{N \times (d+1)} \to \mathbb{R}$ and takes the form below: 
\begin{align}\label{eq:simple_model_for_ntk}
    &f(A) := \left[\sigma\left(\psi\left(  A\right)W^{(1)}\right) W^{(2)}\right]_{N} = \sigma\left(\psi\left(  A\right)_{N,:}W^{(1)}\right)  W^{(2)};
\end{align}
where $\psi:\Real^{N\times(d+1)} \rightarrow\Real^{N\times(d+1)}$ is a feature map (generalizing the attention function defined in~\eqref{eq:psi}), $A \in \mathbb{R}^{(N) \times (d+1)}$ denotes the input data, and $W^{(1)} \in \mathbb{R}^{d+1 \times k}, W^{(2)} \in \mathbb{R}^{k \times 1}$.  

\begin{wrapfigure}{r}{0.45\textwidth}
    \centering
    \includegraphics[width=0.4\textwidth]{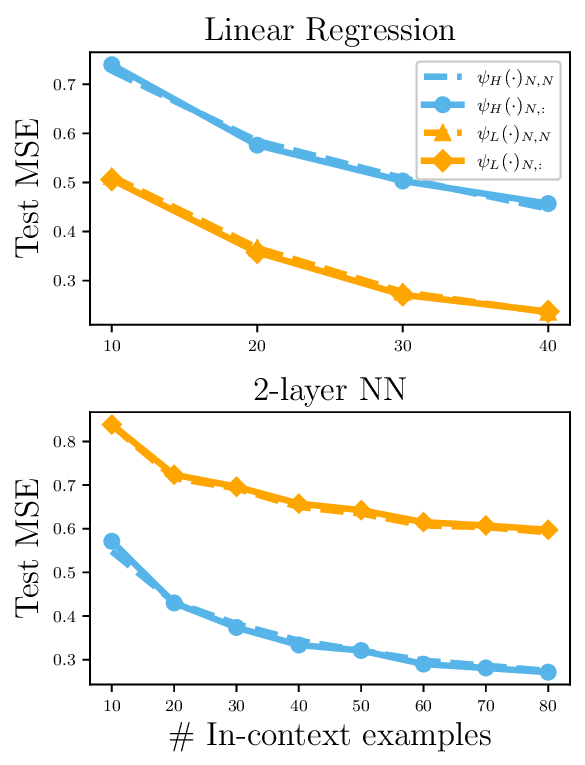}
    \caption{Comparison between using a row of features given by $\psi_K(\cdot)_{N,:}$ and using the scalar given by  $\psi_K(\cdot)_{N,d+1}$ for $K \in \{L, H\}$.  Here, the feature maps $\psi_L, \psi_H$ are defined in~\eqref{eq:one step GD estimate} and~\eqref{eq:hilbert_kernel} respectively.}
    \label{fig:last_element}
    \vspace{2mm}
\end{wrapfigure}

\subsection{Feature-map that enables context-scaling}\label{sub_sec:smoothers}

The key aspect distinguishing the model in \eqref{eq:simple_model_for_ntk} from a standard MLP operating on vectorized inputs $A \in \mathbb{R}^{N (d+1)}$ is the feature map $\psi$. As such, we analyze how the feature map $\psi$ transforms an input
$$A = \begin{bmatrix}
x_1 & x_2 &\cdots&x_{N-1}&x_{N}  \\
y_1 & y_2&\cdots& y_{N-1} & 0\end{bmatrix}^\top \in \Real^{N\times(d+1)}.$$
First, we note that upon varying the function $\psi$, the bottom-right element of $\psi(A)$, denoted $\psi(A)_{N, d+1}$, is capable of implementing several well-known estimators, which we describe below. To ease notation, we let $\mathbf{X} := [x_1,x_2,\cdots,x_N]^\top\in\mathbb{R}^{N\times d}$ and $\mathbf{y}:=[y_1,\cdots,y_N]^\top$. Detailed derivations of the explicit forms for $\psi(A)_{N, d+1}$ below are presented in  Appendix \ref{app:C}.

\textbf{(1) 1-step GD estimate.} Let $\psi_L(A) := (AA^\top)A$.  Then, 
\begin{equation}
\label{eq:one step GD estimate}
\psi_L(A)_{N, d+1} = x_{N}^\top\mathbf{X}^\top\mathbf{y}.
\end{equation}
Thus, $\psi_L$ computes the estimate arising from a linear predictor trained for one step of gradient descent on the data $(\mathbf{X}, \mathbf{y})$.  This estimate has been previously considered as a mechanism through which transformers performed ICL, but there have been no theoretical guarantees for this approach for general ICL tasks beyond linear regression~\citep{von2022transformers,ahn2023transformers,zhang2024trained,mahankali2023one,zhang2024context}.

\textbf{(2) Kernel smoother.} Given a kernel $K$, let $\psi_{K}(A) = \hat{K}(\mathbf{X},\mathbf{X})A$, where 
$$\hat{K}(\mathbf{X},\mathbf{X})_{i, j} = \begin{cases} \frac{K(x_i, x_j)}{\sum_{j\neq i} K(x_i, x_j)} &\text{ if $i \neq j$} \\ 0  &\text{ if $i = j$} \end{cases}.$$
In this case, $\psi_{K}(A)_{N, :}$ has the following form

\begin{align}
    \psi_K(A)_{N,:} = 
    \underbrace{\bigg[\frac{\sum_{i=1}^{N-1} K(x_N, x_i) x_i^\top}{\sum_{i=1}^{N-1} K(x_N, x_i)}}_{\text{smoothed $d$-dimensional features}}, 
    \underbrace{\frac{\sum_{i=1}^{N-1} K(x_N, x_i) y_i}{\sum_{i=1}^{N-1} K(x_N, x_i)}\bigg]}_{\text{smoothed estimate}}  \in \mathbb{R}^{d+1},
\end{align}
and the last element $\psi_{K}(A)_{N, d+1}$ is the kernel smoother estimate,
\begin{equation}
\label{eq: kernel smoother feature map}
\psi_K(A)_{N, d+1} = \frac{\sum_{i=1}^{N-1} K(x_{N}, x_i) y_i}{\sum_{i=1}^{N-1} K(x_{N}, x_i)}.
\end{equation}

Below, we provide key examples of kernel smoothers that can be implemented by~\eqref{eq: kernel smoother feature map} upon changing the kernel $K$.  

\begin{enumerate}
\item When $K$ is the exponential kernel, i.e., $K(z, z') = e^{-z^\top z}$, then $\psi_K$ implements softmax attention, and $\psi_K(A)_{N, d+1}$ is the kernel smoother corresponding to the exponential kernel.

\item When using the kernel $H$ defined in \eqref{eq:hilbert_kernel}, then, $\psi_{H}(A)_{N, d+1}$ implements the Hilbert estimate, which is consistent as the number of in-context examples goes to infinity~\citep{devroye1998hilbert}. 
\end{enumerate}

In our experiments in Figure~\ref{fig:multi_context}, we trained an MLP on features computed using $\psi_K(A)_{N,:}$.  Yet, the results above suggest that the scalar $\psi_{K}(A)_{N, d+1}$ alone should be sufficient for context scaling. Indeed, in the case of the Hilbert estimate, this entry alone provides a consistent estimate as the context length goes to infinity.  To this end, in Figure \ref{fig:last_element}, we compare the performance of two MLPs when the number of tasks is fixed and the context length increases.  The first MLP is trained using $\psi_K(A)_{N, :} \in \mathbb{R}^{d+1}$, and the second is trained on only $\psi_K(A)_{N, d+1}$.  The results in this figure confirm that using $\psi_K(A)_{N, d+1}$ is as good as using $\psi_K(A)_{N, :}$ for context-scaling.  

\subsection{Training MLPs that simultaneously context-scale and task-scale}\label{sub_sec:smoothers_2}

\begin{figure}[t]
    \centering
    \includegraphics[width=1\textwidth]{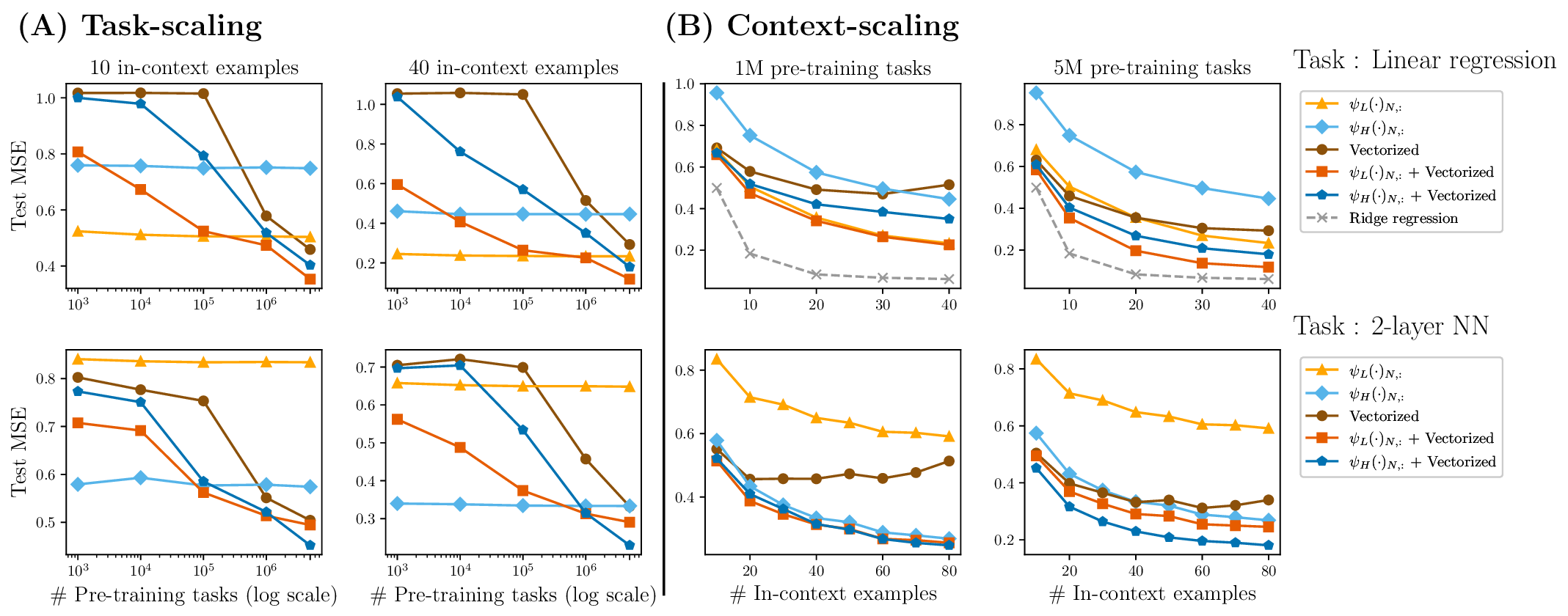}
    \caption{Comparison of MLPs trained using (1) vectorized inputs; (2) features from $\psi_K$ for $K \in L, H$ defined in~\eqref{eq:one step GD estimate} and~\eqref{eq:hilbert_kernel}; (3) both features from $\psi_K$ and vectorized inputs.  We compare performance across two ICL tasks: linear regression and two-layer teacher-student neural networks.
    \textbf{(A)} Task-scaling ability of MLPs using different inputs.
    \textbf{(B)} Context-scaling ability of MLPs using different inputs. MLPs trained on both vectorized inputs and features from $\psi_K$ are able to simultaneously context-scale and task-scale. Experimental details are provided in Appendix \ref{app:A}.
    }
    \label{fig:MLP_ICL}
\end{figure}

As the Hilbert estimate provides a consistent estimate, our results above show that transformers provably generalize to unseen tasks, when the context length approaches infinity.  Nevertheless, the issue with using the Hilbert estimate alone is that the Hilbert estimate is only computed using examples provided in a context.  As such, it cannot task-scale unlike MLPs trained on vectorized inputs.  We now show that training MLPs on vectorized inputs concatenated with features estimated using $\psi_K(A)$ result in MLPs that can both context-scale and task-scale.  

Namely, we revisit the experiment presented in Figure~\ref{fig:weak_icl} and extend our analysis by training MLPs on three distinct input configurations: (1) vectorized input data; (2) features from $\psi_K(A)_{N, :}$; and (3) the concatenation of vectorized input data and features from $\psi_K(A)_{N, :}$. In our experiments, we consider the feature maps $\psi_L$ and $\psi_H$ discussed in the previous section.  The results of training these MLPs is presented in Figure~\ref{fig:MLP_ICL} and we summarize the results below.

\begin{enumerate}
    \item \textbf{MLPs with vectorized input data:} Figure~\ref{fig:MLP_ICL}A demonstrates that these MLPs exhibit task-scaling. Yet, Figure~\ref{fig:MLP_ICL}B reveals that these MLPs fail to context-scale and performance can even deteriorate with increased context length.

    \item \textbf{MLPs with features from $\psi_K(\cdot)_{N, :}$:} Figure~\ref{fig:MLP_ICL}A illustrates that these MLPs do not task-scale, as performance does not improve with an increasing number of tasks.  This behavior matches intuition as the Hilbert smoother and 1-step gradient descent features are task-specific and do not leverage inter-task relationships. Yet, Figure~\ref{fig:MLP_ICL}B shows that these MLPs successfully context-scale, which happens provably for the particular case of $\psi_H$.

    \item \textbf{MLPs with both vectorized inputs and features from $\psi_K(\cdot)_{N, :}$:} Figure~\ref{fig:MLP_ICL}A demonstrates that these MLPs are capable of task-scaling, consistent with the performance of MLPs on vectorized data alone.  Moreover, in Figure~\ref{fig:MLP_ICL}B, we now observe that these MLPs are now capable of context-scaling, consistent with the performance of MLPs using the features from  $\psi_K(\cdot)_{N, :}$ alone.
\end{enumerate}

These results underscore the importance of the feature map $\psi_K$ for context-scaling and highlight the effectiveness of using both vectorized inputs and features from $\psi_K$ in improving the ability of models to learn in-context.

\section{Conclusion}

\paragraph{Summary.} In this work, we observed that transformers, unlike MLPs, are able to simultaneously context-scale (improve performance as the context length increases) and task-scale (improve performance as the number of pre-training tasks increases).  To better understand this property of transformers, we first identified a simplified transformer (SGPT) that could solve ICL tasks competitively with GPT-2 architecture transformers despite having no trainable key, query, and value weights in attention layers.  By studying a one-layer version of SGPT, we identified that the attention operator of SGPT applied a feature map, $\psi$, on input data that enabled context-scaling.  In particular, we showed that this feature map could implement kernel smoothers such as the Hilbert estimate, which is a statistically consistent estimator as the context length goes to infinity.  As such, our work provably demonstrates that transformers can context-scale, generalizing to new, unseen tasks when provided a large context.  We demonstrated the effectiveness of the feature map, $\psi$, for context-scaling by showing that MLPs trained on both features from $\psi$ and vectorized inputs could simultaneously context-scale and task-scale.  

\paragraph{Future work and limitations.}  While we have provably established that one-layer transformers can context-scale, we empirically observe that one-layer transformers are not as sample-efficient as deep transformers for both context-scaling and task-scaling. Thus, an important future direction is understanding how depth improves the sample complexity of transformers in both context-scaling and task-scaling settings. Exploring this aspect remains a promising avenue for future research and could provide a comprehensive understanding of ICL, and more broadly, a better understanding of how transformers are able to generalize to new tasks when provided large contexts.

% \paragraph{Reproducibility Statement.} The codebase is included as supplementary material, and we will release the GitHub repository upon acceptance.

\bibliography{ref}
\bibliographystyle{abbrv}

\appendix
\section{Experiments details}\label{app:A}

We provide all experimental details below.  

\paragraph{Problem formulation.} For all ICL tasks studied in our work, we consider T pretraining tasks, each with input data of the form:
$$A_t = \begin{bmatrix}
x_1^t & x_2^t & \cdots & x_N^t  \\
y_1^t & y_2^t & \cdots & y_N^t
\end{bmatrix}^\top \in \mathbb{R}^{(N)\times(d+1)}$$
where $t \in {1, \ldots, T}$ indexes the tasks, $N$ denotes the maximum context length, and $d$ denotes the input data dimension. 
We define the loss function $L(\theta)$ as:
$$L(\theta; X_1, \ldots, X_T) := \frac{1}{T}\sum_{t=1}^T\left[\frac{1}{N}\sum_{i=1}^{N}(M_{\theta}(X_t^i)-y_{i+1}^t)^2\right],$$
where
$$A_t^i = \begin{bmatrix}
x_1^t & x_2^t & \cdots & x_i^t & x_{i+1}^t \\
y_1^t & y_2^t & \cdots & y_i^t & 0
\end{bmatrix}^\top \in \Real^{(i+1)\times(d+1)},$$
and $M_{\theta}$ denotes the model with trainable parameters $\theta$.

\paragraph{Vectorized input.} By vectorizing input, we mean flattening the input into a vector. After vectorization, $A_t^i$ defined above becomes,

$$A_t^i = \begin{bmatrix}
{x_1^t}^\top & {x_2^t}^\top & \cdots & {x_i^t}^\top & {x_{i+1}^t}^\top&y_1^t & y_2^t & \cdots & y_i^t & 0
\end{bmatrix}^\top\in\Real^{(i+1)(d+1)}.$$

\paragraph{GPT-2.}We used the GPT-2 implementation from prior work \citep{garg2022can,bai2023transformers}, which is based on the Hugging Face implementation~\citep{wolf2020huggingfaces}. Following the approach in these prior works, we modified the embedding layer with a learnable linear layer that maps from the ambient dimension to an embedding dimension. 

\paragraph{SGPT.} To construct SGPT, we make the following modifications to the GPT-2 architecture: (1) we fix all key, query, value weights to the identity; (2) we eliminate all batch-normalization layers; and (3) we remove the second linear layer from each MLP. Following prior work \citep{garg2022can,bai2023transformers}, we modified the embedding layer with a linear layer that maps from the ambient dimension to an embedding dimension.  In SGPT, this linear layer is not trainable and serves as a fixed random map. We outline the architecture below.  

Let $A \in \mathbb{R}^{N \times (d+1)}$ be the input of the model. We initialize a random matrix $W_0 \in \mathbb{R}^{(d+1) \times k}$, where $k$ is the embedding dimension. Defining the input of the $i$-th layer as $H^{(i)}$, we have $H^{(0)} := AW_0$,
\begin{align}
H^{(i)} = \sigma\Big(\big(g(H^{(i-1)})W^{(i)}_{proj} + H^{(i-1)}\big)W^{(i)}_{MLP}\Big) + g(H^{(i-1)})W^{(i)}_{proj} + H^{(i-1)}
\end{align}
where:
\begin{itemize}
    \item $\sigma$ is the activation function, chosen to be GeLU,
    \item $g$ is as defined in Equation \ref{eq:psi}, $\phi$ is row wise $l1$ normalziation.
    \item $W^{(i)}_{proj} \in \mathbb{R}^{k \times k}$ is the projection matrix,
    \item $W^{(i)}_{MLP} \in \mathbb{R}^{k \times k}$ is the MLP weight matrix for the $i$-th layer.
\end{itemize}

The last layer of the network is a  linear layer with weights $W_{O}\in\Real^{k\times 1}$.

\paragraph{MLP architectures.} In all experiments, we use a standard 2-layer ReLU MLP with a width of 1024 units. Given an input vector $x \in \mathbb{R}^{d_{in}}$, the MLP implements a function $f$ of the form
\begin{align}
\label{eq: MLP architecture}
f(x) := \sigma(xW_0)W_1
\end{align}
where $W_0 \in \mathbb{R}^{d_{in} \times 1024}$, $W_1 \in \mathbb{R}^{1024 \times 1}$, and $\sigma$ is the ReLU activation function.

\paragraph{Zero-padding input for MLP.} In all experiments, we always trained \textbf{a single MLP for all context lengths} by zero-padding to the largest context length. For example, if the input data is in $\mathbb{R}^d$ and the largest context length is $N_{max}$, then the input dimension of the MLP is $dN_{max} $.

\paragraph{Expeirmental details for Figure \ref{fig:weak_icl}.} In this experiment, we trained an 8-layer GPT-2 model with 8 attention heads and a width of 256. The MLP configuration is the same as that in~\eqref{eq: MLP architecture}. We trained and tested the models on the same set of context lengths: 5, 10, 20, 30, and 40.

\paragraph{Experimental details for Section \ref{sec:exps}.}

We trained both standard GPT-2 architecture transformers and our proposed SGPT with the following configurations:
\begin{itemize}
    \item Widths: \{256, 512, 1024\},
    \item Number of layers: \{2, 4, 6, 8\},
    \item Number of attention heads for GPT-2: \{2, 4, 8\}.
\end{itemize}

The best performance was achieved with an 8-layer model with a width of 256. For the original GPT-2, the optimal configuration used 8 attention heads.

We outline per-task observations and configurations below:
\begin{enumerate}
    \item \textbf{Linear regression with single noise level:} Following the prior work \cite{garg2022can}, the input dimension is $d=20$ and noise level is $\sigma=0.5$. We trained on context lengths from 10 to 40 with a step size of 5.

    \item\textbf{Linear regression with two noise levels:} Following the prior work of \cite{garg2022can}, the input dimension is $d=20$ and noise levels are $\sigma_1=0.1, \sigma_2=0.5$. We trained on context lengths from 1 to 40 with a step size of 1.
    
    \item\textbf{Decision tree:} Following the prior work of \cite{garg2022can}, the input dimension is $d=20$ with a tree depth of 4. We note that with the input structure \eqref{eq:input_structure}, GPT-2 performs poorly, so in our figure we used the pretrained model from \cite{garg2022can}.
    
    \item\textbf{Two-layer ReLU Neural Networks.} As mentioned before, we chose the width of this family of neural networks to be $r=100$ and the input dimension to be $d=20$. We trained on context lengths from 1 to 100 with a step size of 10. We observed that unlike our model, GPT-2 does not generalize well for context lengths that it has not been trained on.

    \item\textbf{Sparse Linear Regression} As mentioned previously, the ambient dimension of the input is $d=20$, consistent with prior work\citep{garg2022can}, and the effective dimension is $s=3$. We used \texttt{scikit-learn}~\cite{scikit-learn} for the Lasso and Ordinary least sauare performances.
\end{enumerate}

\paragraph{Experimental details for Figure \ref{fig:multi_context}.} In all tasks, we used input dimension $d=8$ following the setting in~\cite{tong2024mlps}. We trained and tested both models on context lengths of 10, 20, 30, 40, and 50.

\paragraph{Experimental details for Figure \ref{fig:last_element}.} We train an MLP (\eqref{eq: MLP architecture}) on  features extracted using $\psi_K(\cdot)_{N, :}$ and linear regression on the scalar $\psi_K(\cdot)_{N, d+1}$.

\paragraph{Experimental details for Figure \ref{fig:MLP_ICL}:} 
\begin{itemize}
    \item \textbf{Linear regression:} We used the same settings as used for the model in~\eqref{eq:attentionhead_gpt2} with $d=8$ and $\sigma=0.22$. We trained and tested models on the context lengths 5, 10, 20, 30, and 40.
    \item \textbf{2-layer NN task:} We used the same settings as used for the model in~\ref{task:nn2layer} with $d=8, r=100$. Trained and tested models on the context lengths 10, 20, 30, 40, 50, 60, 70, and 80. 
\end{itemize}

\paragraph{Hardware.} We used machines equipped with NVIDIA A100 and A40 GPUs, featuring V-RAM capacities of 40GB. These machines also included 8 cores of Intel(R) Xeon(R) Gold 6248 CPU @ 2.50GHz with up to 150 GB of RAM. For all our experiments, we never used more than one GPU, and no model was trained for more than two days.

\section{Feature map derivation}\label{app:C}

Let $A = \begin{bmatrix}
x_1^\top & y_1  \\
x_2^\top & y_2 \\ 
\vdots & \vdots\\
x_{N-1}^\top & y_{N-1}\\ 
x_N^\top & 0\\ 
\end{bmatrix} \in \Real^{N\times(d+1)}$. 

\paragraph{1-step of GD.} In this case, we have
\begin{align*}
    \psi_L(A) & = (AA^\top)A\\
    & = \begin{bmatrix}
    x_1^\top & y_1  \\
    x_2^\top & y_2 \\ 
    \vdots & \vdots\\
    x_{N-1}^\top & y_{N-1}\\
    x_N^\top & 0
    \end{bmatrix}\begin{bmatrix}
x_1 & x_2 &\cdots&x_{N-1}&x_{N}  \\
y_1 & y_2&\cdots& y_{N-1} & 0\end{bmatrix} A \\ 
&=\begin{bmatrix}
    x_1^\top x_1+y_1y_1 & \cdots&x_1^\top x_{N-1}+y_1y_{N-1}&x_1^\top x_{N} \\
    x_2^\top x_1+y_2y_1  & \cdots&x_2^\top x_{N-1}+y_2y_{N-1}&x_2^\top x_{N}  \\ 
    \vdots & \cdots & \vdots & \vdots \\
    x_{N-1}^\top x_1+y_{N-1}y_1 &  \cdots&x_{N-1}^\top x_{N-1}+y_{N-1}y_{N-1}&x_{N-1}^\top x_{N} \\ 
    x_{N}^\top x_1 & \cdots&x_{N}^\top x_N&x_{N}^\top x_{N} 
    \end{bmatrix}\begin{bmatrix}
x_1 & y_1  \\
x_2 & y_2 \\ 
\vdots & \vdots\\
x_{N-1} & y_{N-1}\\ 
x_N & 0\\ 
\end{bmatrix}.
\end{align*}
Thus, $\psi_L(A)_{N, d+1} = x_{N}^\top\mathbf{X}^\top\mathbf{y}$, where $\mathbf{X} := \begin{bmatrix}x_1^\top\\\vdots\\x_N^\top\end{bmatrix}\in\Real^{N\times(d+1)}$ and $\mathbf{y}:=\begin{bmatrix}y_1\\\vdots\\y_N\end{bmatrix}$. This value is equivalent to the prediction given by using one-step of gradient descent to solve linear regression.

\paragraph{Kernel smoothers.} In this case, we have
\begin{align*}
    \psi_K(A) & = \hat{K}(\mathbf{X},\mathbf{X})A\\
&=\begin{bmatrix}
    0 &    \frac{K(x_1,x_2)}{\sum_{\substack{i=1 \ i \neq 1}}^NK(x_1,x_i)}&\cdots&    \frac{K(x_1,x_N)}{\sum_{\substack{i=1 \ i \neq 1}}^NK(x_1,x_i)}\\
    \frac{K(x_2,x_1)}{\sum_{\substack{i=1 \ i \neq 2}}^N K(x_2,x_i)}&0&\cdots&\frac{K(x_2,x_N)}{\sum_{\substack{i=1 \ i \neq 2}}^N K(x_2,x_i)}\\
    \vdots&\vdots&\vdots&\vdots\\
    \frac{K(x_N,x_1)}{\sum_{\substack{i=1 \ i \neq N}}^N K(x_N,x_i)}&\frac{K(x_N,x_2)}{\sum_{\substack{i=1 \ i \neq N}}^N K(x_N,x_i)}&\cdots&0\\
    \end{bmatrix}\begin{bmatrix}
x_1 & y_1  \\
x_2 & y_2 \\ 
\vdots & \vdots\\
x_{N-1} & y_{N-1}\\ 
x_N & 0\\ 
\end{bmatrix}.\\
\end{align*}
Now last row of the $\psi_K(A)$ is given by
\begin{align*}
    \psi_K(A)_{N, :} = \begin{bmatrix}
        \frac{\sum_{i=1}^{N-1}K(X_N,x_i)x_i}{\sum_{i=1}^{N-1}K(X_N,x_i)} &\frac{\sum_{i=1}^{N-1}K(X_N,x_i)y_i}{\sum_{i=1}^{N-1}K(X_N,x_i)}
    \end{bmatrix} \in \Real^{d+1}.
\end{align*}
Thus, $\psi_K(A)_{N, d+1}$ is equvialent to the prediction for $x_N$ given by using a kernel smoother with kernel $K$.

\newpage
\section{Additional experiments}\label{app:D}
% \section{Appendix}\label{app:E}

\begin{figure}[h]
    \centering
    \includegraphics[width=1\textwidth]{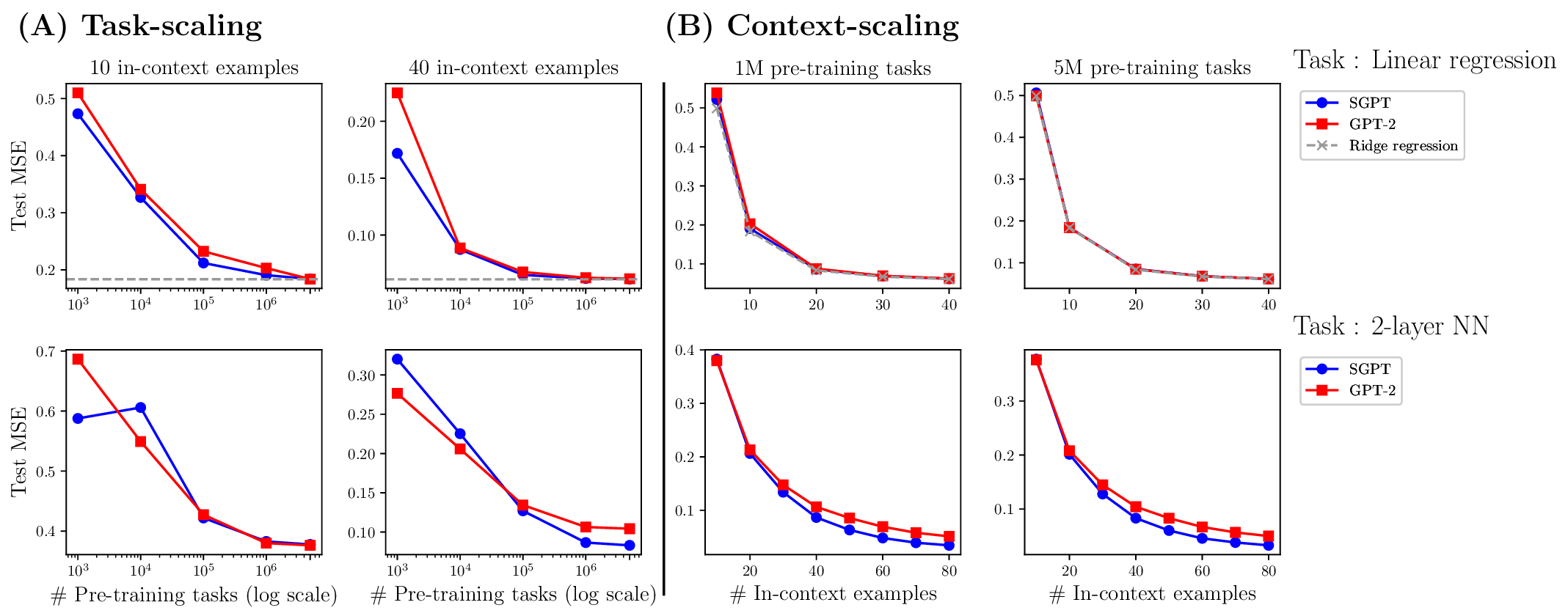}
    \caption{Task-scaling and context-scaling of GPT-2 architecture transformers versus SGPT for linear regression and 2-layer neural networks tasks.  \textbf{(A)} Task-scaling abilities of these models with $10$ in-context examples. \textbf{(B)} Context-scaling abilities of these models with $10^5$ (left) and $10^6$ (right) pre-training tasks. }
    
    \label{fig:SGPT-GPT2}
\end{figure}

In the following experiments, we trained both an 8-layer, 8-head GPT-2 model and an 8-layer SGPT model, using identical context lengths during training and testing. As shown in Figure \ref{fig:SGPT-GPT2}, both models are capable of scale and context scaling. Here for both tasks we set $d=8$ and for the linear regression task $\sigma = 0.22$ and for 2-layer neural network we set the width of the networks to $r=100$.

\end{document}